\documentclass[sigconf]{acmart}
\usepackage{booktabs} % For formal tables
\usepackage{flushend}
\usepackage{amsmath}
\usepackage{mdframed}
\usepackage{soul,xcolor}
\usepackage{float}

\setcopyright{rightsretained}
\acmDOI{10.475/123_4}

\acmISBN{123-4567-24-567/08/06}

\acmConference[DLRS'17]{2nd workshop on Deep Learning for Recommender Systems}{August 2017}{Como, Italy} 
\acmYear{1997}
\copyrightyear{2017}

\acmPrice{15.00}

\begin{document}
\title{Automatic Generation of Natural Language Explanations}

\author{Felipe Costa}
\affiliation{%
  \institution{Aalborg Universitet}
  \streetaddress{Selma Lagerl{\o}fs Vej 300}
  \city{Aalborg}
  \country{Denmark} 
  \postcode{9220}
}
\email{fcosta@cs.aau.dk}

\author{Sixun Ouyang}
\affiliation{%
  \institution{Insight Centre for Data Analytics}
  \institution{University College Dublin}
  \streetaddress{Belfield, Dublin 4, Ireland} 
}
\email{sixun.ouyang@insight-centre.org}

\author{Peter Dolog}
\affiliation{%
  \institution{Aalborg Universitet}
  \streetaddress{Selma Lagerl{\o}fs Vej 300}
  \city{Aalborg}
  \country{Denmark} 
  \postcode{9220}
}
\email{dolog@cs.aau.dk}

\author{Aonghus Lawlor}
\affiliation{%
  \institution{Insight Centre for Data Analytics}
  \institution{University College Dublin}
  \streetaddress{Belfield, Dublin 4, Ireland} 
}
\email{aonghus.lawlor@insight-centre.org}

% The default list of authors is too long for headers}
%\renewcommand{\shortauthors}{B. Trovato et al.}

\begin{abstract}
An important task for recommender system is to generate explanations according to a user's preferences. Most of the current methods for explainable recommendations use structured sentences to provide descriptions along with the recommendations they produce. However, those methods have neglected the review-oriented way of writing a text, even though it is known that these reviews have a strong influence over user's decision. 

In this paper, we propose a method for the automatic generation of natural language explanations, for predicting  how a user would write about an item, based on user ratings from different items' features. We design a character-level recurrent neural network (RNN) model, which generates an item's review explanations using long-short term memories (LSTM). The model generates text reviews given a combination of the review and ratings score that express opinions about different factors or aspects of an item. Our network is trained on a sub-sample from the large real-world dataset BeerAdvocate. Our empirical evaluation using natural language processing metrics shows the generated text's quality is close to a real user written review, identifying negation, misspellings, and domain specific vocabulary.

\end{abstract}

%
% The code below should be generated by the tool at
% http://dl.acm.org/ccs.cfm
% Please copy and paste the code instead of the example below. 
%
\begin{CCSXML}
<ccs2012>
<concept>
<concept_id>10002951.10003317.10003347.10003350</concept_id>
<concept_desc>Information systems~Recommender systems</concept_desc>
<concept_significance>500</concept_significance>
</concept>
<concept>
<concept_id>10010147.10010178.10010179.10010182</concept_id>
<concept_desc>Computing methodologies~Natural language generation</concept_desc>
<concept_significance>500</concept_significance>
</concept>
<concept>
<concept_id>10010147.10010257.10010293.10010294</concept_id>
<concept_desc>Computing methodologies~Neural networks</concept_desc>
<concept_significance>500</concept_significance>
</concept>
</ccs2012>
\end{CCSXML}

\ccsdesc[500]{Information systems~Recommender systems}
\ccsdesc[500]{Computing methodologies~Natural language generation}
\ccsdesc[500]{Computing methodologies~Neural networks}

% We no longer use \terms command
%\terms{Theory}

\keywords{Recommender Systems, Explainability, Explanations, Neural Networks}

%% Used in some conference proceedings e.g. sigplan and sigchi
% \begin{teaserfigure}
%   \includegraphics[width=\textwidth]{sampleteaser}
%   \caption{This is a teaser}
%   \label{fig:teaser}
% \end{teaserfigure}

\maketitle

\section{Introduction}

One of the key challenges for a recommender system is to predict the probability that a target user likes a given item, taking into account the user's history and their similarity to other users. However, making predictions in this way does not explain why the item matches with the users' preferences. Recent works have introduced the concept of explainable recommender systems, which try to generate explanations according to users' preferences rather than only predicting a numerical rating for an item. In this work we develop an approach using character-level neural networks to generate readable explanations. 

Current explainable recommendations propose to mine user reviews to generate explanations. In \cite{Zhang:2014} they propose an explicit factor model, where they first extracts aspects and user opinions by phrase-level sentiment analysis on user generated reviews, then generate both recommendations and \textit{disrecommendations} according to the specific product features and personalised to the user's interests and the hidden features learned. On the other hand, in \cite{He:2015} they propose a  tripartite graph to enrich the user-item binary relation to a user-item-aspect ternary relation. In each of the these work, they propose to extract aspects from reviews to generate explainable recommendations, but they do not consider user opinions and influences from social relations as a source of explanation. In \cite{Ren:2017} they propose the social collaborative viewpoint regression model, which detects viewpoints and uses social relations as a latent variable model. This model is represented as tuples of a concept, topic, and a sentiment label from both user reviews and trusted social relations. 

Explanations generated in this manner lack natural language expressions, since the sentences are generated in a modular way. However, it is well established by \cite{tintarev2007effective} that a good explanation must be clear, and interesting to the target user, since this information has a significant influence on the user's decision. On-line user-generated reviews present clear and interesting information about items, since they describe personal usage experience from users. Furthermore, this source plays an important role on the user side, since he/she tend to trust the opinion of other users \cite{pu2012evaluating, knijnenburg2012explaining, cosley2003seeing}.

Recurrent neural networks (RNNs) have recently demonstrated to show very good performance in natural language generation, since the generating function can be automatically learned from massive text corpora. Due RNNs suffers from gradient vanishing problem, long-short term memory (LSTM) has been applied to the text generation field, and leads to significant improvements on this issue. Another advantage of using LSTM is the ability to keep in memory the long-range dependencies among words and characters. The combination of RNNs with LSTM have shown promising results on such different text datasets as Shakespeare poem, scientific papers, and linux source code generation \cite{DBLP:journals/corr/KarpathyJL15}.

Most natural languages text generation approaches focus on the raw textual content and often neglect their contextual information. This context, such as the specific location, time and sentiment are important factors in the creation of user generated on-line reviews and should not be neglected. Recent research on recommender systems demonstrated improvements achieved by including context \cite{adomavicius2015context}. This paper incorporates this information to enrich the generated sentences with particular contextual features. 

In this paper, we propose a technique for the automatic generation of explanations, based on generative text reviews given a vector of ratings that express opinions about different factors of an item. Our method is based on a character-level LSTM trained on a sub-sample from the large real-world dataset BeerAdvocate. It is divided into three modules: a context encoder, LSTM decoder, and the review generation. The ratings are normalised, then concatenated to the characters to feed the LSTM cells, which can generate characters that are contextualised by the normalised ratings. The generative review module has a weighted generation based on ratings vector as input. The weights learns soft alignments between generated characters and sentiment, where we adaptively compute encoder-side context vectors used to predict the next characters.

Automatic generated review-oriented explanations, are useful for companies and users, who can benefit from helpfulness aspect of the explanations to assess an item recommendation. \cite{bartoli} shows character-level generation has advantages over other techniques such as unsupervised learning of grammar and punctuation, and can be more efficient than word-level generation, since it allows for the prediction and generation of new words and strings.

This paper presents as contributions:
\begin{itemize}
\item A context-aware review generation based on rating scores
\item Generate readable reviews in a human perspective.
\end{itemize}

\section{Problem Formulation}

In this section, we provide the basic definition and preliminaries to generate natural language explanations. Given a set of items $I$, and target user $u$:

\begin{itemize}
\item An item is a product (beer) represented by $i \in I$.
\item Explicit feedback is an action represented by the matrix $X_u: U \times I \to R$, where $u \in U$ is a user, $i \in I$ is an item, and $r \in R$ represents a rating that the user $u$ have been given to item $i$. Considering the each $r$ rating is a vector corresponding to a set of five features {appearance, aroma, palate, taste, and overall.}
\item Reviews are another explicit feedback in text format represented by the matrix $X_a: U \times I \to T$, where $u \in U$ is a user, $i \in I$ is an item, and $t \in T$ represents a review that the user $u$ have been given to item $i$. 
\end{itemize}

\subsection{Problem Statement}
Ratings are attributes to express opinions from a user about a certain item, however it is difficult to compose a judgement of a product based only on the rating score. Therefore, user-generated reviews are richer, since the user can give explanations according to different features and aspects of a specific item. There are many approaches to generating explanations for different types of recommender systems, including collaborative filtering \cite{herlocker2000explaining} and case-based approaches \cite{mcsherry2005explanation}. Explanations showed to increase the effectiveness of the recommendation and the user's satisfaction \cite{tintarev2012evaluating} in various evaluations methods.
%Current recommender systems does not give explanations why an item is recommended, becoming even more difficult to explain why an item is not recommended given other alternatives. Explanations may help users make better decisions in a real world.
Current state of the art in explainable recommender systems does not offer human-oriented explanations. To address this particular issue, our model is defined to target the problem of generating explanations in a review-oriented and natural language basis.

We formulate the item explanation generation problem as follows. Given input ratings vector $r_i =(r_1,\dotso,r_{|r_i|})$, we aim to generate item explanation $e_i = (w_1,\dotso,w_{|t_i|})$, maximizing the conditional probability $p(e|r)$. Note, rating $r_i$ is the average values from the evaluation of target item $i$ in a fixed numerical representation, while the review $t_i$ is considered a character sequence of variable length. We set $|r|$ as $5$ in our task, as we have $5$ features with different ratings values. The model learns to compute the likelihood of generated reviews given a set of input ratings. This conditional probability $p(e|r)$ is represented in the Eq. \ref{eq:first}.

\begin{equation}
\begin{aligned}
p(e|r)=\prod_{s=1}^{|e|} p(w_s|w<s,r)
\end{aligned}
\label{eq:first}
\end{equation}

where $w<s = (w_1,\dotso,w_{t-1})$

\section{Related Works}

Neural networks have started to attract attention in recommender systems community only recently. In \cite{DBLP:conf/acml/KoMG16} they study recurrent neural networks in different architectures for a collaborative recommender system with experiments showing good performance. Despite good performance, this example of work suffers from the same problem as the other works that it is not explainable.

The work of \cite{Almahairi:2015:LDR:2792838.2800192} is among the first where a recommender system is utilising the review text as side information to improve the performance of recommender system and the solutions are rooted in recurrent neural networks. Our work differs from this work as we are in fact trying to generate explanations in the form of a user-generated review to improve a user's understanding of recommended items.

What we would like to achieve however is an alignment between variables or features which lead to a recommendation of one item or another and a descriptive text where rules about the text composition are learned from the existing reviews. Therefore, we would like to achieve similar alignment as others have achieved in different domains such as text generation for images as in \cite{DBLP:journals/pami/KarpathyF17}.

Learning the rules for generating the reviews can be accomplished by representing input as sentences, words or characters. In \cite{Meng2016} and \cite{DBLP:journals/corr/TangYCZM16} they propose a tree-based neural network model for natural-language inference based on words and their context. We study character-level explanation generation to further improve the state of the art. The work of \cite{DBLP:journals/corr/KarpathyJL15} provides the first insights into why the LSTM variant of neural networks has such good performance. Similar technique were used on \cite{bartoli}, where they build on the previous work to generate product reviews in the restaurant domain. 

Encoding rating vectors in the training phase allows the system to calculate the probability of the next character based on the given rating. In previous work, \cite{dong} showed an efficient method for generation of next the word in the sequence when we add an attention mechanism, showing that this idea improves performance for long sequences.

Character-level generation has shown improvement over word-level on the text generation problem using RNNs \cite{bartoli}. This is because, on the character-level, the neural network can autonomously learn grammatical and punctuation rules. In \cite{bartoli} they mention the character-level RNN provides slightly worse performance than the equivalent word-based model, however it shows improvements in terms of computational cost, which grows with the size of the input and output dictionaries, an in contrast, it allows for the prediction and generation of new words and strings. %Applying the attention-based mechanism concatenation the user vector representation, the model tends to approximate the writing style to the target user.

In \cite{DBLP:journals/corr/LiptonVM15} they focus on character-level review generation and classification where the ratings are used as auxiliary information. Our work differs from both aforementioned approaches for character-level text generation in utilising richer data (ratings are used to explicit quality of a product in different features, identified as a source of user's preference) and providing a first attempt to generate explanations with character level networks to reflect user differences and preferences.

\section{Generated Explanations}
\subsection{Recurrent Neural Network}

%Recurrent neural networks are feed forward networks, which contains at least one back-propagation connection, allowing it flows in a loop \cite{elman}. This structure enables the networks to do temporal processing and learn sequences. Given as input the vector $X$, where the weight matrix is represented by $W_H$, the recurrent neural network is represented as a non-linear equation matrix to predict the output sequence $Y_t$, as shown on the Eq.1.

Recurrent neural networks (RNNs) are feed-forward networks with temporal verifying activation, processing and learning sequential data. While in the training step, given an input vector $X_t$ in time $t$ and the cell state of previous time step $t-1$, where the input weight matrix is represented by $W_{x}$ and state weight metrices refers to $W_{h}$, the RNNs then pass the cell state $h_t$ to the next time step and propose a prediction value $Y_t$ via a softmax layer which consists of a non-linear softmax function, as shown in Eq. \ref{eq:second}.

%allowing it flows in a loop \cite{elman}. This structure enables the networks to do temporal processing and learn sequences. Given as input the vector $X$, where the weight matrix is represented by $W_H$, the recurrent neural network is represented as a non-linear equation matrix to predict the output sequence $Y_t$, as shown on the Eq.1.

\begin{equation}
\begin{aligned}
& h_t = \tanh(X_t \odot W_{x} + h_{t-1} \odot W_{h}) \\
& Y_t = \mathrm{softmax}(h_t \odot W + b)
\end{aligned}
\label{eq:second}
\end{equation}

According to Eq. \ref{eq:second}, if we continue feeding the same values to $X_t$, the input weight matrix $W_{x}$ and state weight matrix $W_{h}$ will be changed to suit the input value.
%, where the previous state will covered. 
% This is a vanishing gradient problem that RNNs have always suffered, when depending on the activation functions, sequential information gets lost over the time. To handle this issue, \cite{hochreiter} introduced Long short term memory(LSTM) cells, what was later improved by \cite{gers} using forget gates to discard some information. 
RNN's suffer a vanishing gradient problem, that depending on the activation functions, sequential information gets lost over time. To handle this issue, \cite{hochreiter} introduced Long short term memory (LSTM) cells, and was later improved by \cite{gers} using forget gates to discard some information. 
%However, RNNs suffer from a vanishing gradient problem, when depending on the activation functions, sequential information gets lost over the time. This represents an issue from this structure, since the weights matrices through time are where the information from the past is stored, i.e. if the weight reaches a value of 0 or 1,000,000, the previous state will not give the right probability for the next output. To handle this issue, \cite{hochreiter} introduced LSTM cells, what was later improved by \cite{gers} using forget gates to discard some information. 

LSTM is an improved version of RNNs controlled by sequential connection of gates: forget gate, input gate and output gate. When receiving an input data $x_t$ at time $t$ and the cell state $C_{t-1}$ from previous time step $t-1$, those values will be concatenated together for the next computation. It will feed the forget gate initially, where it decides which information has to be discarded. There, $f_{t}$ represents the results via the forget gate in time $t$, $W_f$ and $b_f$ refers to the weight matrix and bias, respectively. The next step for LSTM cells is to determine which information should be stored in cell state through the input gate. At the update step, $i_t$ means the input gate results, $W_i$ and $b_i$ are its parameters. The cell creates a candidate state $C'_{t}$ through a $tanh$ layer. Using the candidate state with the previous cell state, forget gate results $f_{t}$ and input gate results $i_{t}$ to update the current state $C_{t}$. Finally, the data goes to output gate, where it uses $sigmoidal$ function layer to determine which part of the cell state is the output, then it multiplies $tanh$ with the current cell state $C_t$ to give as result the character with the highest probability.
%which have as input data $X'$ denoting the concatenation of real input and internal state, activated by a $tanh$ or $sigmoid$ function. Each LSTM cell contains three sigmoidal functions \footnote{Sigmoid function is necessary to compute the internal vector with values between 0 and 1} to forget ($f$), update ($u$), and result ($r$). The output of each LSTM layer is another sequence, allowing us to build a network with several layers of LSTMs. At step $t$, each LSTM layer $H_t$ receives input from the previous layer $C_t$, where $C_t$ is an internal state called memory, representing in each step the new contents of LSTM memory will be the previous content subtracted what the network might forget added the updated memory from my new inputs. $H_t$ is the external state, controlling the end of recursion. LSTM is formally defined as Eq. \ref{eq:third}.
\begin{equation}
\begin{aligned}
& X = [x_{t}, C_{t-1}] \\
& f_{t} = \sigma(X \odot W_f + b_f)\\
& i_{t} = \sigma(X \odot W_i + b_i) \\
& C'_{t} = \tanh(X \odot W_c + b_c) \\
& C_{t} = f_{t} \odot C_{t-1} + i_{t} \odot C'_{t} \\
& o_{t} = \sigma(X \odot W_o + b_o) \\
& H_t = o_{t} \odot \tanh(C_t) \\
\end{aligned}
\label{eq:third}
\end{equation}

\subsection{Generative Concatenative Network(GCN)}
\begin{figure}[t]
    \centering
%    \showthe\columnwidth % Use this to determine the width of the figure.
    \includegraphics[width=0.25\textwidth]{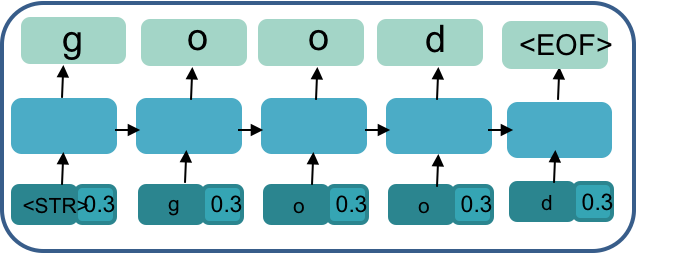}
    \caption{Generative Concatenative Network}
    \label{fig:gcn}
\end{figure}
Generative RNN models can be applied in many fields as most data can be represented as a sequence, especially for text generation. State weights benefits generative RNNs to generate coherent text, where one character can be fed into the network at a time step and these affect the state weights. This project builds on the generative concatenative network presented by \cite{DBLP:journals/corr/LiptonVM15}, which uses an LSTM RNN character-based generation model, adding auxiliary information according to ratings for different feature preferences.
%generate positive or negative sentences. 
%so the time dependent weights are used for what came before in the sequence.

In \cite{DBLP:journals/corr/KarpathyJL15} they define a character-level language model given a sequence of characters as input to an LSTM neural network, calculate the probability of the next character in the sequence with a $softmax$ function at each time step $s$ then generate the character as output. Given a set of $C$ characters we encode all characters with $C-dimensional$ 1-of-C vectors $\{x_t\}, t=1,\dotso,T$, and feed them to the recurrent network to obtain a sequence of $H-dimensional$ hidden vectors as the last layer of the network $\{H_t^l\}, t=1,\dotso,T$. To obtain predictions for the next character in the sequence, the output goes to the top layer of a sigmoid activation function to a sequence of vectors $\hat{y}$, where $\hat{y} = W_y.H^L_t$ and $W_y$ is a $[K \times D]$ parameter matrix. The output vectors are interpreted as holding the $log$ probability of the next character in the sequence and the objective is to minimize the average cross-entropy loss over all targets.

In \cite{DBLP:journals/corr/LiptonVM15} they propose to generate text, conditioned on an auxiliary input $x_{aux}$, where the input $x_{aux}$ is concatenated with the character representation $x^{(t)}_{char}$, as it is seen in Fig. \ref{fig:gcn}. They train their network based on the concatenated information input $x'^{(t)} = [x^{(t)}_{char}; x_{aux}]$. At training time, $x_{aux}$ is a feature of the training set, while during the generation step, they define some $x_{aux}$, concatenating it with each character sampled from $\hat{y}^{(t)}$. They replicate the auxiliary information $x_{aux}$ at each input to allow the model to focus on learning the complex interactions between the auxiliary input and the language, rather than just memorising the input. However, they consider only the overall rating or temperature for a certain item, neglecting the user's preference in different aspects.

\subsection{Context Encoder}

Similar to \cite{DBLP:journals/corr/KarpathyJL15} and \cite{DBLP:journals/corr/LiptonVM15}, our model is based on LSTM RNNs network to generate reviews. Our model adds a set of auxiliary information to each character in the context encoder module.

\begin{figure}
    \centering
    \includegraphics[width=\columnwidth]{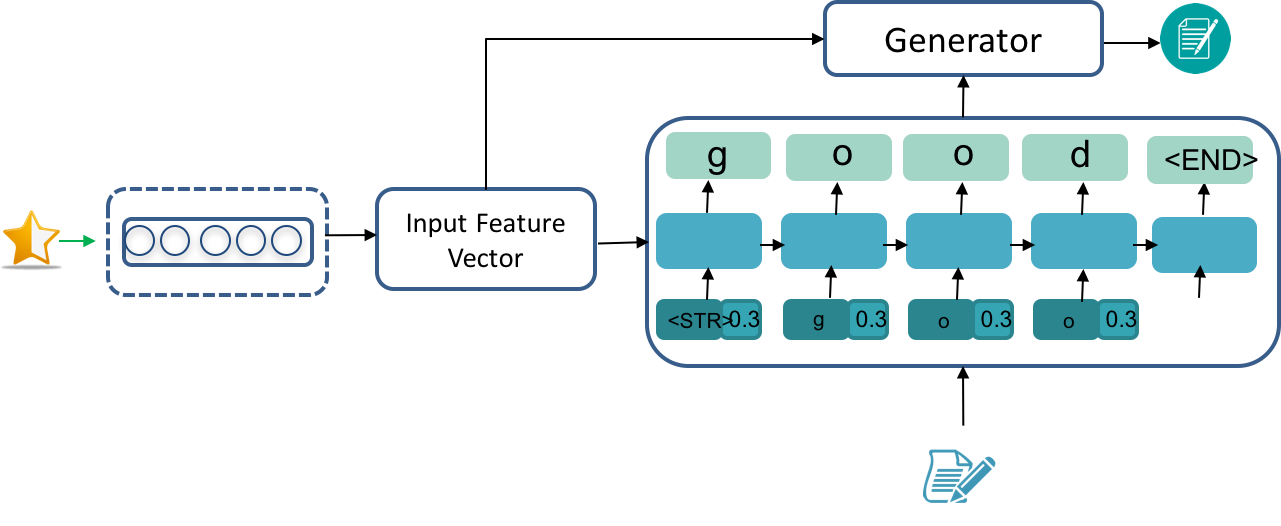}
    \caption{Generative Explanations}
    \label{fig:gcn2}
\end{figure}

In our model, the context encoder module encodes the input character using one-hot encoding and concatenates a set of ratings to it, before feeding it into network as we can see in Fig. \ref{fig:gcn2}. In our experiments, we generate a dictionary for all the characters in the corpus to record their positions, which will be used as the encoding process in the training step and for decoding in the generating step. For each character in the reviews, a one-hot vector will be generated by using its position in that dictionary. Then the one-hot vector will be concatenated with a set of auxiliary informations which relies on the review, as shown in Eq. \ref{eq:context_encoder}. Meanwhile, in terms of the auxiliary information, our model uses a set of numeric values of the users' ratings, which are rescaled to the range $[0, 1]$.

%we propose one hot vector for each text input, i.e. character, concatenating with its ratings, as shown in Eq. \ref{eq:context_encoder}, as the input data to training.

\begin{equation}
\begin{aligned}
X'_{t} = [one hot(x_{char}); x_{auxiliary}]
\label{eq:context_encoder}
\end{aligned}
\end{equation}

%TODO: write this sentence more clearly
% Context encoder module has as input the rating vector, representing the normalized average rating of each item according to different features: appearance, aroma, palate, taste, and overall impression, as shown at Figure \ref{fig:gcn2}.  Then, these vectors are concatenated and fed into a hidden layer which outputs the encoding vector. The result of the hidden layer is computed as shown at Eq. \ref{eq:fifth}.

% \begin{equation}
% \begin{aligned}
% X' = tanh(X \odot [char; r] + b_c)
% \end{aligned}
% \label{eq:fifth}
% \end{equation}

% where $[char; r]$ is concatenated character and ratings  vector, $tanh$ is a non-linear function, $X'  \in  \mathbb{R}^{L \times m}$ is a weight matrix, and $b_c \in  \mathbb{R}^{Ln}$ is the bias. Next, the vector a is used to initialize the n-dimensional hidden vectors of the L-layer recurrent neural networks in the decoder.

\subsection{Generative Explanation}
As mentioned previously, \cite{DBLP:journals/corr/LiptonVM15} proposed a GCN model concatenating characters with some auxiliary information, i.e. overall rating or temperature, being able to generate some remarkable samples. It uses one piece of auxiliary information to enrich the probability to define the next character. 
% TODO: what does this next sentence mean?
%This process has shown an improvement on the quality of training phase. 

%where they encode the decode the characters concatenating with the overall rating, considering it as an auxiliary information which can enrich the probability to define the next character. This process has shown an improvement on the quality of training phase. 

%During the text generation, given an item id, and a rating vector representing the normalized average rating of each item, whom generates different review explanations to the user. We propose an improvement to the concatenation process, where we consider ratings scores for different features on the training and generation phases, instead of only the overall ratings.

We propose an improvement to the concatenation process, where we consider a vector of auxiliary data, i.e a set of the ratings scores for different features of items, instead of only one dimension of auxiliary information. During the review generation our model generates distinct pieces of text tuned to the distribution of applied ratings.

A non-linear $softmax$ layer is used in our model to compute the probability for all characters. During the generation process the model concatenates a $prime$ text, which is a $start$ symbol in each review, concatenated with a series of ratings scores to the model. Then the model passes its output to a $softmax$ layer, as shown in \ref{eq:out_put}, where $H_t$ is the output of a LSTM cell, $W$ and $b$ are the weight and bias of $softmax$ layer, respectively. 
%TODO: what is ???
%Based on the classification, with the ??? defined by a dictionary we generated previously, it generates text by finding the position which has maximum probability.
%we extract the position of the character which has the highest probability, with the help of the dictionary we generated previously to generate a character. 

\begin{equation}
\begin{aligned}
& Y_t = softmax(H_t \odot W + b)
\end{aligned}
\label{eq:out_put}
\end{equation}

This procedure is applied recursively and a group of characters is generated until we find the pre-defined $end$ symbol.
%Recursively, according to this procedure, a group of characters will be generated until it finds the pre-defined $end$ symbol.

%Likewise, we concatenate the given character with the a series of ratings values, feeding into model to get the next generated character. According to this procedure, a group of characters will be generated until they meet the end symbol. This steps shown as \ref{eq:out_put}, where $ h_t $ is the output of LSTM cells 

%where given the personalized generative model, it can produce a predict the next token by maximizing the likelihood of the text, given a classification.

%During the generation process, the model uses the standard softmax function, changing different values of ratings on the range of classifications given by the dataset as: appearance, aroma, palate, taste, and overall impression. Different combinations of the ratings allows the engine to generate personalized reviews to a target user.

By using LSTM cells for character-level explainable review generation, and merging with the vector of ratings, we allow the model to learn grammar and punctuation, being more efficient than word-level models \cite{bartoli}, since our model can predict and generate new words and strings. Therefore, our model generates explanations for recommender systems with a review-oriented perspective, adding improvements on the quality of the explanation text presented to to the user in the form of a review.

% Formalization of the encoding rating vector (in this paper it is referred as context) and use it to predict output characters is shown on the Eq. \ref{eq:sixth}. The equation shows on each time-step $t$ is computed the attention score $s$ of feature $f_i$ as:

% \begin{equation}
% \begin{aligned}
% s_i^t = tanh(X^s \odot [char; r] + b_c) 
% \end{aligned}
% \label{eq:sixth}
% \end{equation}

% where the brackets $[\bullet;\bullet]$ denote concatenation, and $X'  \in  \mathbb{R}^{L \times m}$ is a parameter matrix. Next, the attention context vector $c^t$ is obtained by Eq. \ref{eq:seventh}.

% \begin{equation}
% \begin{aligned}
% c^t = \sum_{i=1}^{|f_i|} c_i^t[char; r]
% \end{aligned}
% \label{eq:seventh}
% \end{equation}

% which is a weighted sum of the ratings and character vectors. We then apply the vector $v^t$ to predict the $t$-th output sentence by Eq. \ref{eq:eighth}.

% \begin{equation}
% \begin{aligned}
% x_t^{att} = tanh(X_1^{c_t} + X_2x_t^L) 
% \end{aligned}
% \label{eq:eighth}
% \end{equation}

\section{Experiments}

\subsection{Parameters Definition}

Empirical experiments used a customised LSTM RNN library written in Python and using Tensorflow. There are 2 hidden layers with 1024 LSTM cells per layer. During training, a wrapper mechanism is used to prevent over-fitting. Feed-in data was split by 100 batches with batch size of 128 and each batch has a sequence length of 280.

\subsection{Dataset}
% has to go here, so it appears on the next page!
\begin{figure*}
\includegraphics[width=0.9\textwidth]{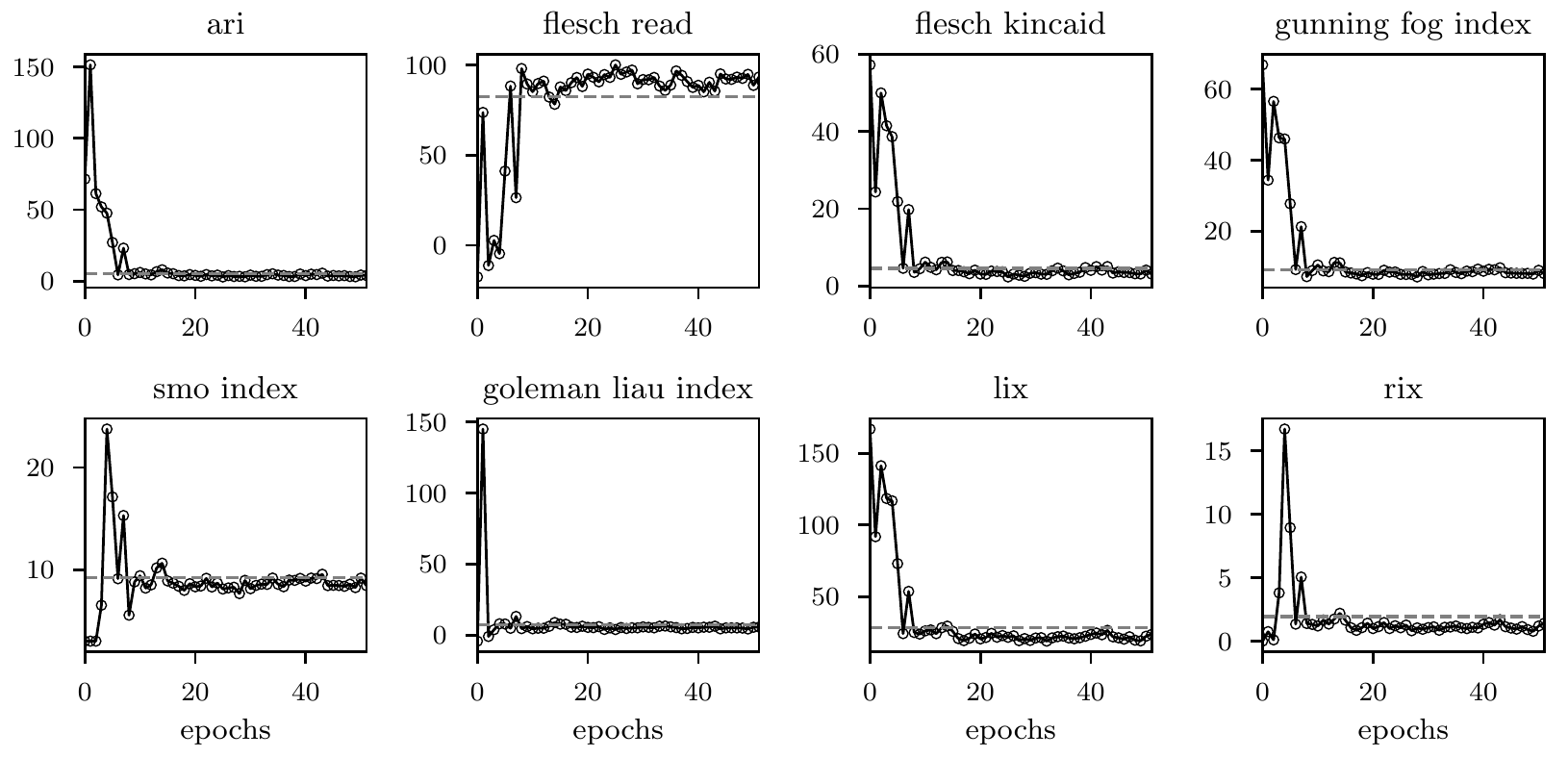}
\caption{Readability metrics with epoch. The metrics are detailed in Sec. \ref{sec:metrics}.}
\label{fig:score_matrix}
\end{figure*}

We tested our model in a sub-sample from the large real-world dataset: BeerAdvocate. The original dataset consists of approximately 1.5 million reviews retrieved from 1998 to 2011. Each review includes rating \footnote{Original ratings were normalized with values between 0-1.} in terms of five categories: \emph{appearance}, \emph{aroma}, \emph{palate}, \emph{taste}, and \emph{overall} impression. Reviews include item and user ids, followed by each of these five ratings, and a plain text review. The summarised statistical information from the extracted sub-sample is shown on the Table \ref{tab:table1}.

\begin{table}[H]
\centering
\begin{tabular}{ll}
\multicolumn{2}{c}{\textbf{BeerAdvocate}}                               \\ \hline
\multicolumn{1}{|l|}{\textbf{\# Users}}   & \multicolumn{1}{l|}{2,815}    \\ \hline
\multicolumn{1}{|l|}{\textbf{\# Beers}}   & \multicolumn{1}{l|}{1,372}    \\ \hline
\multicolumn{1}{|l|}{\textbf{\# Reviews}} & \multicolumn{1}{l|}{4,999} \\ \hline
\end{tabular}
\caption{Dataset Statistics}
\label{tab:table1}
\end{table}

\subsection{Data Preparation}
%In addition to the five ratings categories, generally, there are a set of typical characters that describe the beer in reviews, i.e, appearance, taste, mouthfell, drinkability and smell. In order to depict a clear reviews, a common style among these reviews is using abbreviation of these characters as the start of a description, for instance, "a: very dark with some pretty good lacing as it goes down. s: caramel malt; slightly sweet with some chocolate." where "a" refers to appearance and "s" means the smell of that beer. 

The BeerAdvocate dataset contains several beer categories, and we selected a sub-sample dataset based on just 5 categories: "american ipa" ,"russian imperial stout" ,"american porter","american amber/red ale" and  "fruit/vegetable beer". Considering some reviews are probably too short or even empty that would cause problems with training, we filter our sub-sample to include only reviews with at least 50 characters. For our experiments we concentrate on generating reviews conditioned on the size of reviews of each beer categories, we select 4k reviews of each category for our training datasets.

We first generate a dictionary for all characters, i.e. punctuation, numbers and letters, then transform each character into a one-hot vector using that dictionary. We train the network based on a sequential approach, where each review is fed into a sequence, to do so it is essential to remind the network of the $start$ and $end$ position of each of the reviews. 
%Meanwhile, because it is sequence based model, a prime sequence should be used for generate review sample. 
We do this by appending $start$ and $end$ symbols, i.e. $<\#str\#>$ and $<\#end\#>$, to each reviews for both the training and generation modules. In order to generate explanations for different ratings, we concatenate the input characters with the ratings of the review the character belongs to. In addition, we normalise the scale of the ratings to $[0,1]$.

\subsection{Evaluation Metrics}
\label{sec:metrics}
Current methods to explain recommendations do not have a natural language way to present the information to the user. Our proposed method explains the recommendation to a target user in a style of a user-generated review. To measure the quality of the presented text, we used a suite of natural language readability metrics : Automated Readability index (ARI) \cite{liu2015}, Flesch reading ease (FRE) \cite{Pera1}, Flesch-Kincaid grade level (FGL) \cite{Pera2}, Gunning-Fog index (GFI) \cite{Pera1}, simple measure of gobbledygook (SMOG) \cite{smog}, Coleman Liau index (CLI) \cite{Pera2}, LIX \cite{lix}, and RIX\cite{Pera2}.
%producing an approximate representation of the US grade level needed to understand the text as
Flesch reading ease score is considered the oldest method to calculate the readability through the analysis of number of words and sentence length. An updated version of this metric is the Flesch-Kincaid grade level. %
%, although they use the same core measures, they have different weighting factors. It is important to note the scores between both metrics correlate approximately inversely: a text with a comparatively high score on the reading ease test should have a lower score on the grade level test.
The Gunning Fog index is commonly used to confirm a text can be read easily by the intended audience. %
%, with scores until 12 are considered the best, since scores less than 8 presents a general understanding text and until 12 the text has a good structure for a wide audience. 
% However, this metric has limitations, since it considers words with more than three syllables as complex words. In this context, even words as "interesting", which is not difficult to read, is considered as complex. A short word can be difficult if it is not used very often by most people. The frequency with which words are in normal use affects the readability of text. 
The SMOG score is a improvement of Gunning Fog index, showing better accuracy overall. %
%To calculate a text's readability as simple as possible an approximate formula was also given - count the words of three or more syllables in three 10-sentence samples, estimate the  square root, and add 3.
Automated Readability index relies on a relation of the number of characters per word. % presenting a faster to calculate, as the number of characters is more readily and accurately counted by computer programs than syllables. It considers in its formula the characters as the number of letters, words is the number of spaces, and sentences is the number of sentences. Their score varies from 1 to 14, where the highest score presents a readable text from a college level.
% The Coleman-Liau index relies on characters instead of syllables per word, as ARI metric. They justify their formula saying the characters are more readable and accurately counted by computer programs than are syllables. Unlike syllable-based readability indices, it does not require that the character content of words be analyzed, only their length in characters. Therefore, it could be used in conjunction with theoretically simple mechanical scanners that would only need to recognize character, word, and sentence boundaries.
%LIX score rather than a count of syllables, polysyllabic words, or unfamiliar words as judged by a word list, 
The Lix score gauges the word length by the percentage of long words. %Gauging words makes Lix more objective and quicker to compute than other formulas. The sentence factor adopted (sentence length) is the average number of words per sentence, as in the Flesch. Finally, word and sentence factors are weighted equally, and this, too, contributes to ease of calculation. Improvement was made to this formula, originating the RIX which has better performance for higher amount of text.

\subsection{Results}
\begin{figure}[b!]
\includegraphics[width=\columnwidth]{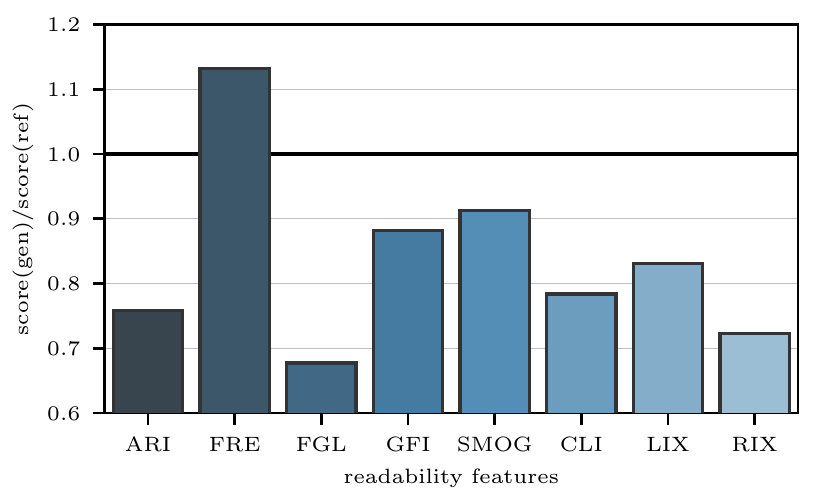}
\caption{Extent of readability for different metrics. The readability of generated text is shown relative the mean score for the user-generated reviews.}
\label{fig:score_compare}
\end{figure}
The initial test of our explanation generation is about readability. We use 8 readability evaluation metrics as mentioned above for both generated and reference reviews.

We first select 10 reviews from our sample dataset as the reference reviews. By using the same users and items from these 10 reviews, as well as considering the different learning curve of the model in different epochs, we generated 10 reviews per epoch from the model. We then apply the readability metrics to the generated and reference reviews to evaluate the text. 
%Furthermore, using mean manipulation on all reviews, we can get the average readability scores for generated and reference reviews. 
The readability results are shown in Fig. \ref{fig:score_matrix}, where it is observed the generated reviews reach the same level of readability as the user reviews on all metrics after 20 epochs.
%where the solid line refers to the average evaluations scores of generated reviews per epochs and the imaginary line represents the average evaluation scores of reference reviews.

As Fig. \ref{fig:score_matrix} shows, the readability evaluation metrics illustrate the capacity of the model to generate reviews which are close to the user's style of writing. 
%According to the time step epochs, the readability scores of generated reviews get closer to the score of reference reviews. Considering reference reviews is written by human, which should be readable, the two scores closer, the more readable the generated review will be.
%Despite the 8 sub-figure of Fig. \ref{fig:score_matrix} demonstrates, the generated reviews is closer to reference reviews ultimately, in terms of overall readability scores. 
We use the readability scores of the generated reviews from the final epochs and normalise to the scores obtained from the reference reviews to demonstrate the relative readability in Figure \ref{fig:score_compare}. This emphasises the neural network generated reviews are close in style to the human written reviews. This is determined by a broad range of readability metrics which are sensitive to different qualities of the text. It is important for our explanations that they are legible, easy to understand, and appear to be written in a recognisable style.
%If generated reviews is as same readable as reference reviews, that calculation would be $1$. More precisely, the closer the extent to $1$, the generated reviews are likely human writing. The extent of readability reveals in Fig. \ref{fig:score_compare}, where $y$ axis is the extent of readability and $x$ axis is the abbreviation of the 8 readability evaluation metrics.
%that "ari" means automated readability index, "fr" refers to flesch reading ease, "fk" represents flesch-kincaid grade level, "gf" reveals gunning fog index, "smo" equals to simple measure of gobbledygook, "lix" stand for LIX and "rix" refers to RIX.

% As Fig. \ref{fig:score_compare} demonstrates, despite the extend on flesch-kincaid grade level is over 0.6 but lower than 0.7, others extend scores are all over 0.7, which reveals the generated reviews is not like machine writing or randomly generating, but likewise human writing. Moreover, as the value of flesch reading ease, gunning fog index, simple measure of gobbledygook and LIX suggests, in terms of human writing, the generated reviews are over 80\% similar to human writing.

% Despite the readability evaluation, we evaluate the generated reviews on different ratings, to investigate the effect of auxiliary values on generation. Due the auxiliary values reveals the opinion of a user for a beer, we used their average ratings values to represent each users preference and general rating opinion to a target beer. By calculating cosine similarity on those ratings values between each user and beer, we identified their dissimilarities. 
We established our model can generate natural language text which reaches the overall readability level of the user-generated reviews. We now investigate the different kinds of explanations that can be generated when we modify the auxiliary values at the generation stage. We are using the ratings from 5 aspects of the beers as auxiliary values, and they represent each users preferences and general ratings opinion about a target beer. We choose a user-item pair $(U, I)$, and compute the average ratings for each feature for both user $\bar{R}_{user} = \sum_{i}R_{user,i}/|R_{user}|$ and item $\bar{R}_{beer}$. The precise contribution of user/item ratings is controlled with a weighting parameter $\alpha$, and we demonstrate three different text samples to compare through Eq. \ref{eq:personal}. As Eq. \ref{eq:personal} shows, $\alpha$ controls the auxiliary values and we then generate reviews based on them. When $\alpha$ is close to $1$, the generated review will be more like a review that the user will write. With $\alpha$ close to 0, the generated review will be closer to the general rating of all users for that beer. To investigate the divergence of generated reviews, we set $\alpha$ equal to {1, 0.5, 0}, which refers to the opinion of the user on general beers, the review the user might compose on that beer, and the general reviews of the beer.
\begin{figure}[h]%
\noindent%
\raggedright{$\alpha=1.0$}\\\fbox{\begin{minipage}[t]{0.9\columnwidth}%
a: pours a clear amber color with a small white head. the aroma is a little spicey \hl{\textbf{with some sweetness. the taste is also a bit more like a beer. there's a little bit of a sweet fruitiness to it as well. the mouthfeel is a bit thin for the style. drinkability is good. i would drink this all night long, but i wouldn't try to get more than one.}}
\end{minipage}}\vspace{2mm}\hfill%
\raggedright{$\alpha=0.5$}\\\fbox{\begin{minipage}[t]{0.9\columnwidth}%
a: pours a clear amber color with a small white head that dissipates quickly.  smell: a little spice. taste is also a bit like a belgian wine. there's a little bit of a banana flavor to it. it is a little too sweet and a little strange.  mouthfeel: thin and watery. drinkability: i would drink this beer again, \hl{\textbf{but i wouldn't try to get my entire offering into a pint glass.   i was pretty surprised this is the beer i was expecting a bit more.}}
\end{minipage}}\vspace{2mm}\hfill%%
\raggedright{$\alpha=0.0$}\\\fbox{\begin{minipage}[t]{0.9\columnwidth}%
a: pours a clear amber color with a small white head that dissipates quickly.  smell: a little spice. taste is also a bit like a belgian wine. there's a little bit of a banana flavor to it. it is a little too sweet and a little strange.  mouthfeel: thin and watery. drinkability: i would drink this beer again, \hl{\textbf{but i wouldn't try to get my entire offerings. i wouldn't recommend it.}}
\end{minipage}}%
\caption{Sample generated reviews for $\alpha=\{1,0.5,0\}$. The dissimilar sentences are highlighted in bold.}
\label{fig:sample_reviews}
\end{figure}
\begin{equation}
\begin{aligned}
R_{auxiliary} = \alpha \times \overline{R}_{user} + (1-\alpha) \times \overline{R}_{beer}
\label{eq:personal}
\end{aligned}
\end{equation} 

According to Fig. \ref{fig:sample_reviews}, the first review ($\alpha=1$) shows the opinion of the user on general beers, which have a positive sentiment overall. When we look into the last review ($\alpha=0$), the common view of that beer is in a negative sentiment, represented by negative sentence, as "I wouldn't recommend it". With $\alpha=0.5$ the generated review displays a relatively neutral attitude towards the beer.

\section{Conclusions}
In this paper, we propose a model to automatically generate natural language explanations for recommender systems. 

%In addition to the beer reviews, the users also provide ratings that express opinions on different factors of an beer. In our specific scenario of beer recommendation, we experimented and evaluated a sub-sample from the large real-world dataset, BeerAdvocate, on the series of beer features, i.e. appearance, aroma, palate, taste, and overall rating.

%had as features: appearance, aroma, palate, taste, and overall evaluation from the large real-world dataset BeerAdvocate sample. 

%Our model is divided in three modules: a context encoder, LSTM decoder, and the review generation. We first encoded the characters using one hot encoding, then concatenate normalised ratings to the characters vectors, defining the characters vector as $[one hot(char); ratings]$. LSTM cells are fed by that with the previous cell state, allowing it to learn the contextual relationship from the characters vectors. The generative review module, feeding by weighted ratings, composes reviews in a natural language manner. These weights allows the network to have soft alignments between generated the characters and sentiment.

Our explanations provide easily intelligible and useful reasons for a user to decide whether to purchase a certain product. This has important benefits for the field of recommender systems since these explanations can help a user to make a better decision and more quickly, as users place a high degree of trust in the reviews of others.

As our experiments with natural language readability metrics show, we were able to generate readable English text with specific characteristics that match user-generated review text. 

In the future we will focus on further extensions of the automatic generation of natural language explanations in two ways: (1) personalised explanations that benefit the user's preferences, where the explanation of the product is tailored to the users ratings, preferred aspects and expressed sentiments; (2) we will test our model in larger reviews domains such as hotels and restaurants.

%\end{document}  % This is where a 'short' article might terminate

\begin{acks}
This work is supported by Science Foundation Ireland through through the Insight Centre for Data Analytics under grant number SFI/12/RC/2289, and Conselho Nacional de Desenvolvimento Cient\'{\i}fico e Tecnol\'ogico - CNPq (grant\# 206065/2014-0).
\end{acks}

\bibliographystyle{ACM-Reference-Format}
\bibliography{sigproc} 

%%% -*-BibTeX-*-
%%% Do NOT edit. File created by BibTeX with style
%%% ACM-Reference-Format-Journals [18-Jan-2012].

\begin{thebibliography}{00}

%%% ====================================================================
%%% NOTE TO THE USER: you can override these defaults by providing
%%% customized versions of any of these macros before the \bibliography
%%% command.  Each of them MUST provide its own final punctuation,
%%% except for \shownote{}, \showDOI{}, and \showURL{}.  The latter two
%%% do not use final punctuation, in order to avoid confusing it with
%%% the Web address.
%%%
%%% To suppress output of a particular field, define its macro to expand
%%% to an empty string, or better, \unskip, like this:
%%%
%%% \newcommand{\showDOI}[1]{\unskip}   % LaTeX syntax
%%%
%%% \def \showDOI #1{\unskip}           % plain TeX syntax
%%%
%%% ====================================================================

\ifx \showCODEN    \undefined \def \showCODEN     #1{\unskip}     \fi
\ifx \showDOI      \undefined \def \showDOI       #1{#1}\fi
\ifx \showISBNx    \undefined \def \showISBNx     #1{\unskip}     \fi
\ifx \showISBNxiii \undefined \def \showISBNxiii  #1{\unskip}     \fi
\ifx \showISSN     \undefined \def \showISSN      #1{\unskip}     \fi
\ifx \showLCCN     \undefined \def \showLCCN      #1{\unskip}     \fi
\ifx \shownote     \undefined \def \shownote      #1{#1}          \fi
\ifx \showarticletitle \undefined \def \showarticletitle #1{#1}   \fi
\ifx \showURL      \undefined \def \showURL       {\relax}        \fi
% The following commands are used for tagged output and should be
% invisible to TeX
\providecommand\bibfield[2]{#2}
\providecommand\bibinfo[2]{#2}
\providecommand\natexlab[1]{#1}
\providecommand\showeprint[2][]{arXiv:#2}

\bibitem[\protect\citeauthoryear{Adomavicius and Tuzhilin}{Adomavicius and
  Tuzhilin}{2015}]%
        {adomavicius2015context}
\bibfield{author}{\bibinfo{person}{Gediminas Adomavicius} {and}
  \bibinfo{person}{Alexander Tuzhilin}.} \bibinfo{year}{2015}\natexlab{}.
\newblock \showarticletitle{Context-aware recommender systems}.
\newblock In \bibinfo{booktitle}{{\em Recommender systems handbook}}.
  \bibinfo{publisher}{Springer}, \bibinfo{pages}{191--226}.
\newblock


\bibitem[\protect\citeauthoryear{Almahairi, Kastner, Cho, and
  Courville}{Almahairi et~al\mbox{.}}{2015}]%
        {Almahairi:2015:LDR:2792838.2800192}
\bibfield{author}{\bibinfo{person}{Amjad Almahairi}, \bibinfo{person}{Kyle
  Kastner}, \bibinfo{person}{Kyunghyun Cho}, {and} \bibinfo{person}{Aaron
  Courville}.} \bibinfo{year}{2015}\natexlab{}.
\newblock \showarticletitle{Learning Distributed Representations from Reviews
  for Collaborative Filtering}. In \bibinfo{booktitle}{{\em Proceedings of the
  9th ACM Conference on Recommender Systems}} {\em (\bibinfo{series}{RecSys
  '15})}. \bibinfo{publisher}{ACM}, \bibinfo{address}{New York, NY, USA},
  \bibinfo{pages}{147--154}.
\newblock
\showISBNx{978-1-4503-3692-5}


\bibitem[\protect\citeauthoryear{Anderson}{Anderson}{1983}]%
        {lix}
\bibfield{author}{\bibinfo{person}{Jonathan Anderson}.}
  \bibinfo{year}{1983}\natexlab{}.
\newblock \showarticletitle{Lix and rix: Variations on a little-known
  readability index}.
\newblock \bibinfo{journal}{{\em Journal of Reading\/}} \bibinfo{volume}{26},
  \bibinfo{number}{6} (\bibinfo{year}{1983}), \bibinfo{pages}{490--496}.
\newblock


\bibitem[\protect\citeauthoryear{Bartoli, d.~Lorenzo, Medvet, Morello, and
  Tarlao}{Bartoli et~al\mbox{.}}{2016}]%
        {bartoli}
\bibfield{author}{\bibinfo{person}{A. Bartoli}, \bibinfo{person}{A. d.
  Lorenzo}, \bibinfo{person}{E. Medvet}, \bibinfo{person}{D. Morello}, {and}
  \bibinfo{person}{F. Tarlao}.} \bibinfo{year}{2016}\natexlab{}.
\newblock \showarticletitle{''Best Dinner Ever!!!'': Automatic Generation of
  Restaurant Reviews with LSTM-RNN}. In \bibinfo{booktitle}{{\em 2016
  IEEE/WIC/ACM International Conference on Web Intelligence (WI)}}.
  \bibinfo{pages}{721--724}.
\newblock


\bibitem[\protect\citeauthoryear{Cosley, Lam, Albert, Konstan, and
  Riedl}{Cosley et~al\mbox{.}}{2003}]%
        {cosley2003seeing}
\bibfield{author}{\bibinfo{person}{Dan Cosley}, \bibinfo{person}{Shyong~K Lam},
  \bibinfo{person}{Istvan Albert}, \bibinfo{person}{Joseph~A Konstan}, {and}
  \bibinfo{person}{John Riedl}.} \bibinfo{year}{2003}\natexlab{}.
\newblock \showarticletitle{Is seeing believing?: how recommender system
  interfaces affect users' opinions}. In \bibinfo{booktitle}{{\em Proceedings
  of the SIGCHI conference on Human factors in computing systems}}. ACM,
  \bibinfo{pages}{585--592}.
\newblock


\bibitem[\protect\citeauthoryear{Dong, Huang, Wei, Lapata, Zhou, and XuT}{Dong
  et~al\mbox{.}}{}]%
        {dong}
\bibfield{author}{\bibinfo{person}{Li Dong}, \bibinfo{person}{Shaohan Huang},
  \bibinfo{person}{Furu Wei}, \bibinfo{person}{Mirella Lapata},
  \bibinfo{person}{Ming Zhou}, {and} \bibinfo{person}{Ke XuT}.}
\newblock \showarticletitle{Learning to Generate Product Reviews from
  Attributes}.
\newblock


\bibitem[\protect\citeauthoryear{Gers, Schmidhuber, and Cummins}{Gers
  et~al\mbox{.}}{2000}]%
        {gers}
\bibfield{author}{\bibinfo{person}{Felix~A Gers}, \bibinfo{person}{J{\"u}rgen
  Schmidhuber}, {and} \bibinfo{person}{Fred Cummins}.}
  \bibinfo{year}{2000}\natexlab{}.
\newblock \showarticletitle{Learning to forget: Continual prediction with
  LSTM}.
\newblock \bibinfo{journal}{{\em Neural computation\/}} \bibinfo{volume}{12},
  \bibinfo{number}{10} (\bibinfo{year}{2000}), \bibinfo{pages}{2451--2471}.
\newblock


\bibitem[\protect\citeauthoryear{He, Chen, Kan, and Chen}{He
  et~al\mbox{.}}{2015}]%
        {He:2015}
\bibfield{author}{\bibinfo{person}{Xiangnan He}, \bibinfo{person}{Tao Chen},
  \bibinfo{person}{Min-Yen Kan}, {and} \bibinfo{person}{Xiao Chen}.}
  \bibinfo{year}{2015}\natexlab{}.
\newblock \showarticletitle{TriRank: Review-aware Explainable Recommendation by
  Modeling Aspects}. In \bibinfo{booktitle}{{\em Proceedings of the 24th ACM
  International on Conference on Information and Knowledge Management}} {\em
  (\bibinfo{series}{CIKM '15})}. \bibinfo{pages}{1661--1670}.
\newblock
\showISBNx{978-1-4503-3794-6}


\bibitem[\protect\citeauthoryear{Herlocker, Konstan, and Riedl}{Herlocker
  et~al\mbox{.}}{2000}]%
        {herlocker2000explaining}
\bibfield{author}{\bibinfo{person}{Jonathan~L Herlocker},
  \bibinfo{person}{Joseph~A Konstan}, {and} \bibinfo{person}{John Riedl}.}
  \bibinfo{year}{2000}\natexlab{}.
\newblock \showarticletitle{Explaining collaborative filtering
  recommendations}. In \bibinfo{booktitle}{{\em Proceedings of the 2000 ACM
  conference on Computer supported cooperative work}}. ACM,
  \bibinfo{pages}{241--250}.
\newblock


\bibitem[\protect\citeauthoryear{Hochreiter and Schmidhuber}{Hochreiter and
  Schmidhuber}{1997}]%
        {hochreiter}
\bibfield{author}{\bibinfo{person}{Sepp Hochreiter} {and}
  \bibinfo{person}{J{\"u}rgen Schmidhuber}.} \bibinfo{year}{1997}\natexlab{}.
\newblock \showarticletitle{Long short-term memory}.
\newblock \bibinfo{journal}{{\em Neural computation\/}} \bibinfo{volume}{9},
  \bibinfo{number}{8} (\bibinfo{year}{1997}), \bibinfo{pages}{1735--1780}.
\newblock


\bibitem[\protect\citeauthoryear{Karpathy and Fei{-}Fei}{Karpathy and
  Fei{-}Fei}{2017}]%
        {DBLP:journals/pami/KarpathyF17}
\bibfield{author}{\bibinfo{person}{Andrej Karpathy} {and} \bibinfo{person}{Li
  Fei{-}Fei}.} \bibinfo{year}{2017}\natexlab{}.
\newblock \showarticletitle{Deep Visual-Semantic Alignments for Generating
  Image Descriptions}.
\newblock \bibinfo{journal}{{\em {IEEE} Trans. Pattern Anal. Mach. Intell.\/}}
  \bibinfo{volume}{39}, \bibinfo{number}{4} (\bibinfo{year}{2017}),
  \bibinfo{pages}{664--676}.
\newblock


\bibitem[\protect\citeauthoryear{Karpathy, Johnson, and Li}{Karpathy
  et~al\mbox{.}}{2015}]%
        {DBLP:journals/corr/KarpathyJL15}
\bibfield{author}{\bibinfo{person}{Andrej Karpathy}, \bibinfo{person}{Justin
  Johnson}, {and} \bibinfo{person}{Fei{-}Fei Li}.}
  \bibinfo{year}{2015}\natexlab{}.
\newblock \showarticletitle{Visualizing and Understanding Recurrent Networks}.
\newblock \bibinfo{journal}{{\em CoRR\/}}  \bibinfo{volume}{abs/1506.02078}
  (\bibinfo{year}{2015}).
\newblock
\newblock
\shownote{International Conference on Learning Representaions.}


\bibitem[\protect\citeauthoryear{Knijnenburg, Willemsen, Gantner, Soncu, and
  Newell}{Knijnenburg et~al\mbox{.}}{2012}]%
        {knijnenburg2012explaining}
\bibfield{author}{\bibinfo{person}{Bart~P Knijnenburg},
  \bibinfo{person}{Martijn~C Willemsen}, \bibinfo{person}{Zeno Gantner},
  \bibinfo{person}{Hakan Soncu}, {and} \bibinfo{person}{Chris Newell}.}
  \bibinfo{year}{2012}\natexlab{}.
\newblock \showarticletitle{Explaining the user experience of recommender
  systems}.
\newblock \bibinfo{journal}{{\em User Modeling and User-Adapted Interaction\/}}
  \bibinfo{volume}{22}, \bibinfo{number}{4-5} (\bibinfo{year}{2012}),
  \bibinfo{pages}{441--504}.
\newblock


\bibitem[\protect\citeauthoryear{Ko, Maystre, and Grossglauser}{Ko
  et~al\mbox{.}}{2016}]%
        {DBLP:conf/acml/KoMG16}
\bibfield{author}{\bibinfo{person}{Young{-}Jun Ko}, \bibinfo{person}{Lucas
  Maystre}, {and} \bibinfo{person}{Matthias Grossglauser}.}
  \bibinfo{year}{2016}\natexlab{}.
\newblock \showarticletitle{Collaborative Recurrent Neural Networks for Dynamic
  Recommender Systems}. In \bibinfo{booktitle}{{\em Proceedings of The 8th
  Asian Conference on Machine Learning, {ACML} 2016, Hamilton, New Zealand,
  November 16-18, 2016.}} {\em (\bibinfo{series}{{JMLR} Workshop and Conference
  Proceedings})}, \bibfield{editor}{\bibinfo{person}{Robert~J. Durrant} {and}
  \bibinfo{person}{Kee{-}Eung Kim}} (Eds.), Vol.~\bibinfo{volume}{63}.
  \bibinfo{publisher}{JMLR.org}, \bibinfo{pages}{366--381}.
\newblock


\bibitem[\protect\citeauthoryear{Lipton, Vikram, and McAuley}{Lipton
  et~al\mbox{.}}{2015}]%
        {DBLP:journals/corr/LiptonVM15}
\bibfield{author}{\bibinfo{person}{Zachary~Chase Lipton},
  \bibinfo{person}{Sharad Vikram}, {and} \bibinfo{person}{Julian McAuley}.}
  \bibinfo{year}{2015}\natexlab{}.
\newblock \showarticletitle{Capturing Meaning in Product Reviews with
  Character-Level Generative Text Models}.
\newblock \bibinfo{journal}{{\em CoRR\/}}  \bibinfo{volume}{abs/1511.03683}
  (\bibinfo{year}{2015}).
\newblock


\bibitem[\protect\citeauthoryear{Liu, Koutrika, and Wu}{Liu
  et~al\mbox{.}}{2015}]%
        {liu2015}
\bibfield{author}{\bibinfo{person}{Lei Liu}, \bibinfo{person}{Georgia
  Koutrika}, {and} \bibinfo{person}{Shanchan Wu}.}
  \bibinfo{year}{2015}\natexlab{}.
\newblock \showarticletitle{Learningassistant: A novel learning resource
  recommendation system}. In \bibinfo{booktitle}{{\em Data Engineering (ICDE),
  2015 IEEE 31st International Conference on}}. IEEE,
  \bibinfo{pages}{1424--1427}.
\newblock


\bibitem[\protect\citeauthoryear{Mc~Laughlin}{Mc~Laughlin}{1969}]%
        {smog}
\bibfield{author}{\bibinfo{person}{G~Harry Mc~Laughlin}.}
  \bibinfo{year}{1969}\natexlab{}.
\newblock \showarticletitle{SMOG grading-a new readability formula}.
\newblock \bibinfo{journal}{{\em Journal of reading\/}} \bibinfo{volume}{12},
  \bibinfo{number}{8} (\bibinfo{year}{1969}), \bibinfo{pages}{639--646}.
\newblock


\bibitem[\protect\citeauthoryear{McSherry}{McSherry}{2005}]%
        {mcsherry2005explanation}
\bibfield{author}{\bibinfo{person}{David McSherry}.}
  \bibinfo{year}{2005}\natexlab{}.
\newblock \showarticletitle{Explanation in recommender systems}.
\newblock \bibinfo{journal}{{\em Artificial Intelligence Review\/}}
  \bibinfo{volume}{24}, \bibinfo{number}{2} (\bibinfo{year}{2005}),
  \bibinfo{pages}{179--197}.
\newblock


\bibitem[\protect\citeauthoryear{Meng, Mou, Li, and Jin}{Meng
  et~al\mbox{.}}{2016}]%
        {Meng2016}
\bibfield{author}{\bibinfo{person}{Zhao Meng}, \bibinfo{person}{Lili Mou},
  \bibinfo{person}{Ge Li}, {and} \bibinfo{person}{Zhi Jin}.}
  \bibinfo{year}{2016}\natexlab{}.
\newblock \bibinfo{booktitle}{{\em Context-Aware Tree-Based Convolutional
  Neural Networks for Natural Language Inference}}.
\newblock \bibinfo{publisher}{Springer International Publishing},
  \bibinfo{address}{Cham}, \bibinfo{pages}{515--526}.
\newblock
\showISBNx{978-3-319-47650-6}


\bibitem[\protect\citeauthoryear{Pera and Ng}{Pera and Ng}{2012}]%
        {Pera1}
\bibfield{author}{\bibinfo{person}{Maria~Soledad Pera} {and}
  \bibinfo{person}{Yiu-Kai Ng}.} \bibinfo{year}{2012}\natexlab{}.
\newblock \showarticletitle{BReK12: A Book Recommender for K-12 Users}. In
  \bibinfo{booktitle}{{\em Proceedings of the 35th International ACM SIGIR
  Conference on Research and Development in Information Retrieval}} {\em
  (\bibinfo{series}{SIGIR '12})}. \bibinfo{pages}{1037--1038}.
\newblock
\showISBNx{978-1-4503-1472-5}


\bibitem[\protect\citeauthoryear{Pera and Ng}{Pera and Ng}{2013}]%
        {Pera2}
\bibfield{author}{\bibinfo{person}{Maria~Soledad Pera} {and}
  \bibinfo{person}{Yiu-Kai Ng}.} \bibinfo{year}{2013}\natexlab{}.
\newblock \showarticletitle{What to Read Next?: Making Personalized Book
  Recommendations for K-12 Users}. In \bibinfo{booktitle}{{\em Proceedings of
  the 7th ACM Conference on Recommender Systems}} {\em (\bibinfo{series}{RecSys
  '13})}. \bibinfo{pages}{113--120}.
\newblock
\showISBNx{978-1-4503-2409-0}


\bibitem[\protect\citeauthoryear{Pu, Chen, and Hu}{Pu et~al\mbox{.}}{2012}]%
        {pu2012evaluating}
\bibfield{author}{\bibinfo{person}{Pearl Pu}, \bibinfo{person}{Li Chen}, {and}
  \bibinfo{person}{Rong Hu}.} \bibinfo{year}{2012}\natexlab{}.
\newblock \showarticletitle{Evaluating recommender systems from the user's
  perspective: survey of the state of the art}.
\newblock \bibinfo{journal}{{\em User Modeling and User-Adapted Interaction\/}}
  \bibinfo{volume}{22}, \bibinfo{number}{4} (\bibinfo{year}{2012}),
  \bibinfo{pages}{317--355}.
\newblock


\bibitem[\protect\citeauthoryear{Ren, Liang, Li, Wang, and de~Rijke}{Ren
  et~al\mbox{.}}{2017}]%
        {Ren:2017}
\bibfield{author}{\bibinfo{person}{Zhaochun Ren}, \bibinfo{person}{Shangsong
  Liang}, \bibinfo{person}{Piji Li}, \bibinfo{person}{Shuaiqiang Wang}, {and}
  \bibinfo{person}{Maarten de Rijke}.} \bibinfo{year}{2017}\natexlab{}.
\newblock \showarticletitle{Social Collaborative Viewpoint Regression with
  Explainable Recommendations}. In \bibinfo{booktitle}{{\em Proceedings of the
  Tenth ACM International Conference on Web Search and Data Mining}} {\em
  (\bibinfo{series}{WSDM '17})}. \bibinfo{pages}{485--494}.
\newblock
\showISBNx{978-1-4503-4675-7}


\bibitem[\protect\citeauthoryear{Tang, Yang, Carton, Zhang, and Mei}{Tang
  et~al\mbox{.}}{2016}]%
        {DBLP:journals/corr/TangYCZM16}
\bibfield{author}{\bibinfo{person}{Jian Tang}, \bibinfo{person}{Yifan Yang},
  \bibinfo{person}{Samuel Carton}, \bibinfo{person}{Ming Zhang}, {and}
  \bibinfo{person}{Qiaozhu Mei}.} \bibinfo{year}{2016}\natexlab{}.
\newblock \showarticletitle{Context-aware Natural Language Generation with
  Recurrent Neural Networks}.
\newblock \bibinfo{journal}{{\em CoRR\/}}  \bibinfo{volume}{abs/1611.09900}
  (\bibinfo{year}{2016}).
\newblock


\bibitem[\protect\citeauthoryear{Tintarev and Masthoff}{Tintarev and
  Masthoff}{2007}]%
        {tintarev2007effective}
\bibfield{author}{\bibinfo{person}{Nava Tintarev} {and} \bibinfo{person}{Judith
  Masthoff}.} \bibinfo{year}{2007}\natexlab{}.
\newblock \showarticletitle{Effective explanations of recommendations:
  user-centered design}. In \bibinfo{booktitle}{{\em Proceedings of the 2007
  ACM conference on Recommender systems}}. ACM, \bibinfo{pages}{153--156}.
\newblock


\bibitem[\protect\citeauthoryear{Tintarev and Masthoff}{Tintarev and
  Masthoff}{2012}]%
        {tintarev2012evaluating}
\bibfield{author}{\bibinfo{person}{Nava Tintarev} {and} \bibinfo{person}{Judith
  Masthoff}.} \bibinfo{year}{2012}\natexlab{}.
\newblock \showarticletitle{Evaluating the effectiveness of explanations for
  recommender systems}.
\newblock \bibinfo{journal}{{\em User Modeling and User-Adapted Interaction\/}}
  \bibinfo{volume}{22}, \bibinfo{number}{4} (\bibinfo{year}{2012}),
  \bibinfo{pages}{399--439}.
\newblock


\bibitem[\protect\citeauthoryear{Zhang, Lai, Zhang, Zhang, Liu, and Ma}{Zhang
  et~al\mbox{.}}{2014}]%
        {Zhang:2014}
\bibfield{author}{\bibinfo{person}{Yongfeng Zhang}, \bibinfo{person}{Guokun
  Lai}, \bibinfo{person}{Min Zhang}, \bibinfo{person}{Yi Zhang},
  \bibinfo{person}{Yiqun Liu}, {and} \bibinfo{person}{Shaoping Ma}.}
  \bibinfo{year}{2014}\natexlab{}.
\newblock \showarticletitle{Explicit Factor Models for Explainable
  Recommendation Based on Phrase-level Sentiment Analysis}. In
  \bibinfo{booktitle}{{\em Proceedings of the 37th International ACM SIGIR
  Conference on Research \&\#38; Development in Information Retrieval}} {\em
  (\bibinfo{series}{SIGIR '14})}. \bibinfo{pages}{83--92}.
\newblock
\showISBNx{978-1-4503-2257-7}


\end{thebibliography}

\end{document}